\documentclass[letterpaper, 10 pt, conference]{ieeeconf}
\usepackage{cite}
\usepackage{graphicx}
\usepackage{subfigure}
\usepackage{algorithmicx}
\usepackage{algorithm}
\usepackage[noend]{algpseudocode}
\usepackage{amsfonts}
\usepackage{amsmath}
\usepackage{autobreak}
\usepackage{mathtools}
\usepackage{cite}
\usepackage{flushend}
\usepackage{color}
\usepackage{multirow}
\usepackage{caption}
\usepackage{threeparttable}
\usepackage{booktabs}
\usepackage{hyperref} 

\IEEEoverridecommandlockouts                              

\title{\LARGE \bf
Moving Forward in Formation: A Decentralized Hierarchical Learning Approach to Multi-Agent Moving Together 
}

\author{Shanqi~Liu$^{1}$, Licheng~Wen$^{1}$, Jinhao~Cui$^{1}$, Xuemeng~Yang$^{1}$, Junjie~Cao$^{1}$, and Yong~Liu$^{1,*}$
\thanks {$^{1}$Shanqi~Liu, Licheng~Wen, Jinhao~Cui, Xuemeng~Yang, Junjie~Cao and Yong~Liu are with the State Key Laboratory of Industrial Control Technology and Institute of Cyber-Systems and Control, Zhejiang University, Zhejiang, 310027, China (*Yong Liu is the corresponding author, {\tt\small yongliu@iipc.zju.edu.cn})}
}

\begin{document}

\maketitle
\thispagestyle{empty}
\pagestyle{empty}
\begin{abstract}
        Multi-agent path finding in formation has many potential real-world applications like mobile warehouse robots. However, previous multi-agent path finding (MAPF) methods hardly take formation into consideration. Furthermore, they are usually centralized planners and require the whole state of the environment. Other decentralized partially observable approaches to MAPF are reinforcement learning (RL) methods. However, these RL methods encounter difficulties when learning path finding and formation problem at the same time. In this paper, we propose a novel decentralized partially observable RL algorithm that uses a hierarchical structure to decompose the multi-objective task into unrelated ones. It also calculates a theoretical weight that makes every task’s reward has equal influence on the final RL value function. Additionally, we introduce a communication method that helps agents cooperate with each other. Experiments in simulation show that our method outperforms other end-to-end RL methods and our method can naturally scale to large world sizes where centralized planner struggles. We also deploy and validate our method in a real-world scenario.
\end{abstract}

\section{INTRODUCTION}
Mobile robots have been deployed in many real-world applications nowadays, including drone swarm, aircraft-towing vehicles and warehouse robots\cite{wareroom}\cite{manyrobot}.
In many of these scenarios, it is important for the agents to move in a specific formation while avoiding obstacles \cite{ma2017feasibility}. For example, the warehouse robots need to work together to transport large cargo. However, most current multi-agent path finding (MAPF) algorithms can not plan in such cases as they do not take formation into consideration.

There are few works focused on solving the multi-agent path finding in formation (MAiF) problem\cite{li2020moving}. Most of them are centralized algorithms.
A centralized planner needs the information and intentions of all agents to generate collision-free paths \cite{boyarski2015icbs}. It becomes infeasible when the number of agents grows and the map size enlarges \cite{mellinger2012mixed}. Besides, we believe that considering a partially-observable world is an essential step towards real-world deployment. Therefore, we focus on decentralized methods that rely on a limited field of view to solve MAiF problem.

Most of the former decentralized partially observable methods to MAPF are reinforcement learning (RL) algorithms\cite{primal,wang2020mobile}. They have shown great potential to learn a general policy using local observation\cite{liu2020mapper}. However, learning different tasks together can be challenging, especially when optimizing conflicting objectives. The majority of multi-objective reinforcement learning (MORL) approaches are single-policy algorithms. They adopt a linear scalarization function in order to learn Pareto optimal solutions \cite{ching1979multiple}. The linear scalarization function is a weighted sum of the parameters that the transform multi-objective rewards vector into a single value. However, the weights used during learning relies on manual design and fine-tuning\cite{kusari2020predicting}.

In this paper, we propose a novel hierarchical reinforcement learning algorithm to generalize previous MAPF methods.
The major contributions of this paper are summarized as follow:
\begin{itemize}
        \item We propose a hierarchical reinforcement learning structure to divide the multi-objective task into unrelated ones. We train each task's policy separately by optimizing its own reward.
        \item We propose a novel way to calculate the linear scalarization function based on the well-trained policies.  Our method can calculate the theoretical weights that make each task's reward have equal influence on the final RL value function.
        \item We introduce a communication method that takes up little bandwidth to help agents cooperate with each other.
        \item  We perform extensive experiments in both simulation and real-world scenarios, where our method outperforms other comparison methods and scales to large world sizes where centralized planners struggle.
\end{itemize}
\begin{figure}[tb]
        \centering
        \subfigure[WeTech Robot]{
                \includegraphics[width=0.42\linewidth, height=25mm]{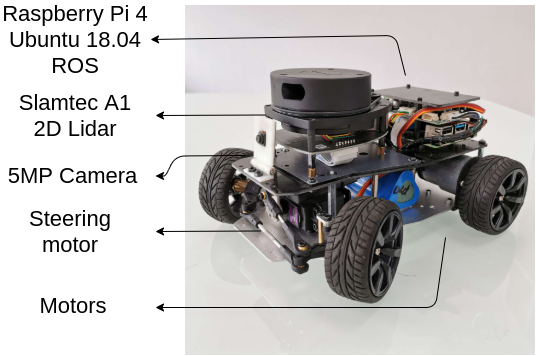}
                \label{nanocar}
        }
        \subfigure[Snapshot of the field test]{
                \includegraphics[width=0.45\linewidth, height=24.7mm]{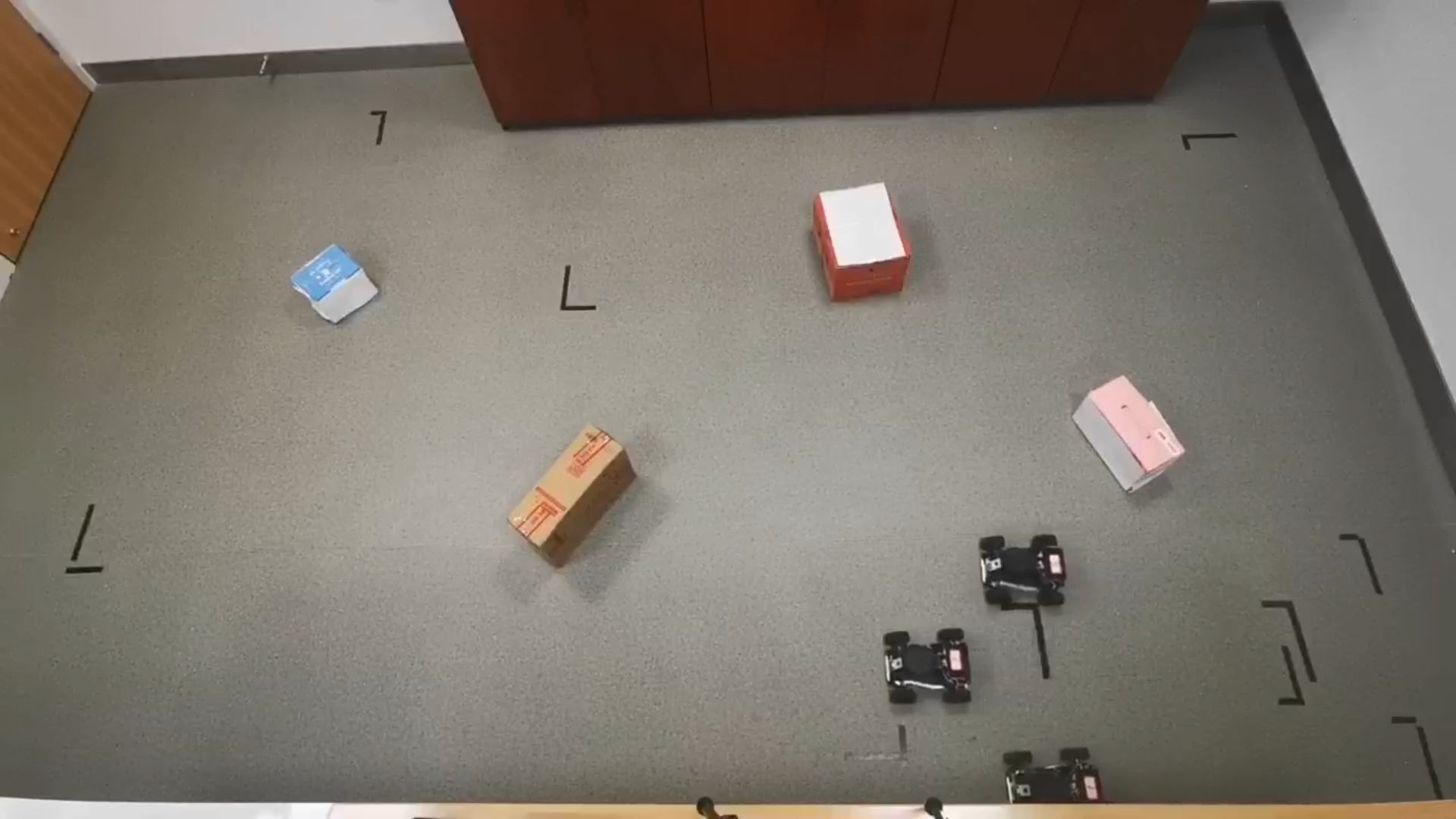}
                \label{experiment}
        }
        \caption{We test the proposed method on ackermann-steering robots produced by WeTech. The experiment video is available in the attachment of this paper.}
        \label{acker}
\end{figure}


\section{RELATED WORKS}
\subsubsection{Moving Agents in Formation (MAiF)}

MAiF, a variant of Multi-Agent Path Finding (MAPF) problem, contains two key sub-tasks: planning collision-free paths for multiple agents while keeping the agents in formation. The former subtask can be addressed by numerous MAPF planner, including reduction-based methods\cite{yu2013planning,surynek2015reduced}, A*-based methods\cite{wagner2011m,standley2010finding,ferner2013odrm}, and dedicated search-based methods\cite{sharon2015conflict,boyarski2015icbs,felner2017search}. Formation-control algorithms can apply to the later sub-tasks. A method of planning motion for formations of non-holonomic robots is presented in \cite{barfoot2004motion}, and a control method for a team of mobile robots maintaining and changing the desired formation using graph theory is presented in \cite{desai2001modeling}. The MAiF problem is formally proposed in \cite{li2020moving}. In this work, the authors develop a two-phase search algorithm to solve both sub-tasks simultaneously.

\subsubsection{Multi-Agent Reinforcement Learning (MARL)}
The most important problem encountered when training a multi-agent policy is the curse of dimensionality. Most centralized approaches fail as the combination of state-action spaces is an exponential explosion, requiring impractical amounts of training data to converge. Thus, many works have focused on centralized training and decentralized executing(CTDE) policy learning. VDN\cite{vdn}, QMIX\cite{qmix} and Qtran\cite{qtran} represent Q learning based CTDE methods. We use VDN to train our policy in this work. There are other methods like MADDPG\cite{maddpg} and COMA\cite{coma} base on actor-critic structure to train a CTDE policy.
\subsubsection{Hierarchical Reinforcement Learning (HRL)}
There are several RL approaches to learning hierarchical policies\cite{sutton1999between,kulkarni2016hierarchical,bakker2004hierarchical}. However, these have many strict limits and are not off-policy training methods. Recently popular works like HIRO\cite{nachum2018data}, Option-Critic\cite{bacon2017option} and FeUdal Networks\cite{vezhnevets2017feudal}have achieve quite good performance. Especially, HAC\cite{levy2017learning} using hindsight\cite{andrychowicz2017hindsight} to overcome non-stationary that came from continually changing sub-policy in HRL so it can use off-policy method to have a better performance. However, in our work, we can directly use off-policy method as we already had a well-trained sub-policy.
\subsubsection{Multi-Objective Reinforcement Learning (MORL)}
MORL algorithms have two main categories\cite{morlgeneralized,morlmulti}: single-policy methods and multiple-policy methods. Single-policy approaches usually aim to find the optimal policy for a given weight among the objectives\cite{morlmanaging,morlsteering}. Multi-policy methods aim to learn a set of policies to obtain the approximate Pareto frontier. They usually perform multiple runs of a single-policy method over different preferences\cite{morlprediction,morl2015multi,morlmanifold}.

\section{POLICY REPRESENTATION}
\subsection{Observation and Basic Action Definition}

\begin{figure}[tbp]
        \centering
        \includegraphics[width=0.8\linewidth]{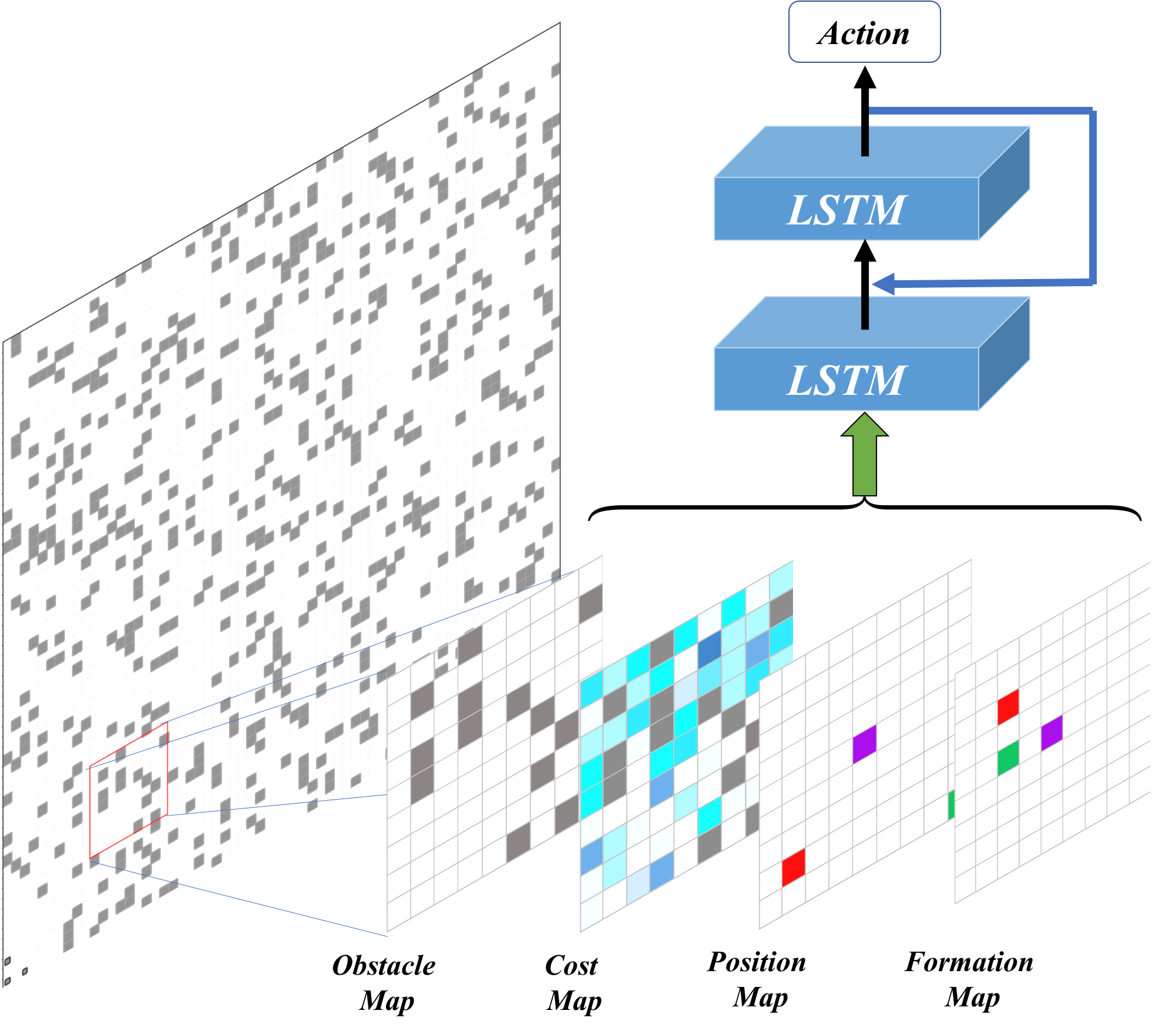}
        \caption{Observation space of simulation environment and our model structure.}
        \label{obs}
\end{figure}

We consider a partially observable discrete grid graph, where agents can only observe the state of the world in a limited field of view around themselves (9x9 FOV in practice). As shown in Fig. \ref{obs}, our observation are divided into four channels:\textit{i)} Obstacle map: the obstacle grids equal 1, empty grids equal 0; \textit{ii)} Position map: the grid contains other agents equals the id of that agent, otherwise zero, different color stands for different agents in figure; \textit{iii)} Cost map: the cost of the shortest path from each grid to the agent's goal, this observation is pre-computed before training, different color stands for different cost in figure; \textit{iv)} Formation map: it contains the desired formation of all agents.

Agents take five discrete actions in the grid world: moving to one of the four cardinal cells or staying still. At each time step, some actions may be invalid, such as moving into a wall.
\subsection{Formation Loss Function}
Inspired by Procrustes Analysis\cite{gower1975generalized}, we define our loss function between two formations. We assume $k$ agents in Cartesian plane has two formations $X_1=((x_1,y_1),...(x_k,y_k))$ and $X_2=((w_1,z_1),...(w_k,z_k))$. We define the formation loss $L_{f}$ between $X_1$ and $X_2$ as:
\begin{equation}
        L_{f}\left(X_{1}, X_{2}\right)=\left\|X_{2}- X_{1} \Gamma-1_{k} \gamma^{\mathrm{T}}\right\|^{2}
\end{equation}
where
\begin{equation*}
        \begin{array}{c}
                \Gamma=\mathbf{M}(\theta)  ,
                \theta=\tan ^{-1}\left(\frac{\sum_{i=1}^{k}\left(w_{i} y_{i}-z_{i} x_{i}\right)}{\sum_{i=1}^{k}\left(w_{i} x_{i}+z_{i} y_{i}\right)}\right)             \vspace{1ex} \\
                \gamma=\left[\frac{\sum w_{i}-\sum x_{i}}{K}, \frac{\sum z_{i}-\sum y_{i}}{K}\right]
        \end{array}
\end{equation*}
The $\|X\|=\left\{\operatorname{trace}\left(X^{\mathrm{T}} X\right)\right\}^{1 / 2}$ is the Euclidean norm, and $\mathbf{M}(\theta)$ denotes a 2D rotation matrix with angle $\theta$. The loss function we designed is more reasonable than the function in \cite{li2020moving} for they didn't consider the rotation transformation between agents' formations.

\subsection{Policy Definition}

We have three policies used in this work: path finding policy, formation policy and meta policy. Each policy has its unique task, which will be introduced below. The overall hierarchical reinforcement learning structure will be discussed in \ref{hrl}.
\subsubsection{Path Finding Policy}
Path finding policy aims at solving the MAPF problem without considering the formation loss. In order to reduce the training difficulty, we propose an action clipping method to reduce the action dimensions.
The original MAPF problem usually contains swapping conflicts defined as agents planning to swap locations in a single time step. It only occurs when two agents have the opposite goal directions, which is impossible in the MAiF scenario because all agents in the formation ought to move toward the same goal direction.
As all other collisions can be solved by one agent waiting when the other agent moves towards the goal position, we can optimize path finding policy by clipping the non-optimal actions. The non-optimal action is defined as the action, which makes the value increase in the cost map.
We define a reward that encourages agents to find the path in Table. \ref{table1}.

\subsubsection{Formation Policy}
The formation policy focus on keeping agents in a specific formation.
We design a training environment, which randomly generates agents and keeps all agents within at least one agent's field of view.  We also clip the action space by forbidding invalid moves(run into walls). The reward can be seen in Table. \ref{table1}. $L_{f}$ in the table is the formation loss.
\subsubsection{Meta Policy}
Meta policy is a high-level policy that decides which low-level policy should be used at each step. The reward is stated in Table. \ref{table1}. The $w_f$ is the basic weight of $L_{f}$ that uses to balance the reward of path finding and keep formation. More details will be discussed in Section. \ref{morl}.

\begin{table}[h]
        \centering
        \caption{ALL REWARD STRUCTURES}
        \begin{tabular}{cccc}
                \hline
                                      & Path finding & Formation & Meta policy       \\
                \hline
                Agent Collision       & -50          & -50       & -50               \\
                Movement Towards Goal & 1            & 0         & 1                 \\
                No Movement           & -0.25        & -0.25     & 0                 \\
                Finish Episode        & 100          & 0         & 0                 \\
                Formation Loss        & 0            & $L_{f}$   & $w_f \cdot L_{f}$ \\
                Keep Formation        & 0            & 100       & 0                 \\
                \hline
        \end{tabular}
        \begin{tablenotes}
                \item[1] All rewards are valued in each time step.
        \end{tablenotes}
        \label{table1}
\end{table}

\section{METHOD}
In our paper, we propose a novel hierarchical reinforcement learning method that can decompose the path finding and formation problem into unrelated ones. Comparing to end-to-end reinforcement learning methods, our method can significantly reduce the learning difficulty and easily transfer to new situations.
Furthermore, we propose a novel approach to calculate the theoretical linear scalarization function, which can balance the path finding and formation policies. Finally, we introduce a communication method that benefits the training process of decentralized cooperative policy.

\begin{figure*}[t]
        \centering
        \includegraphics[width=0.8\textwidth]{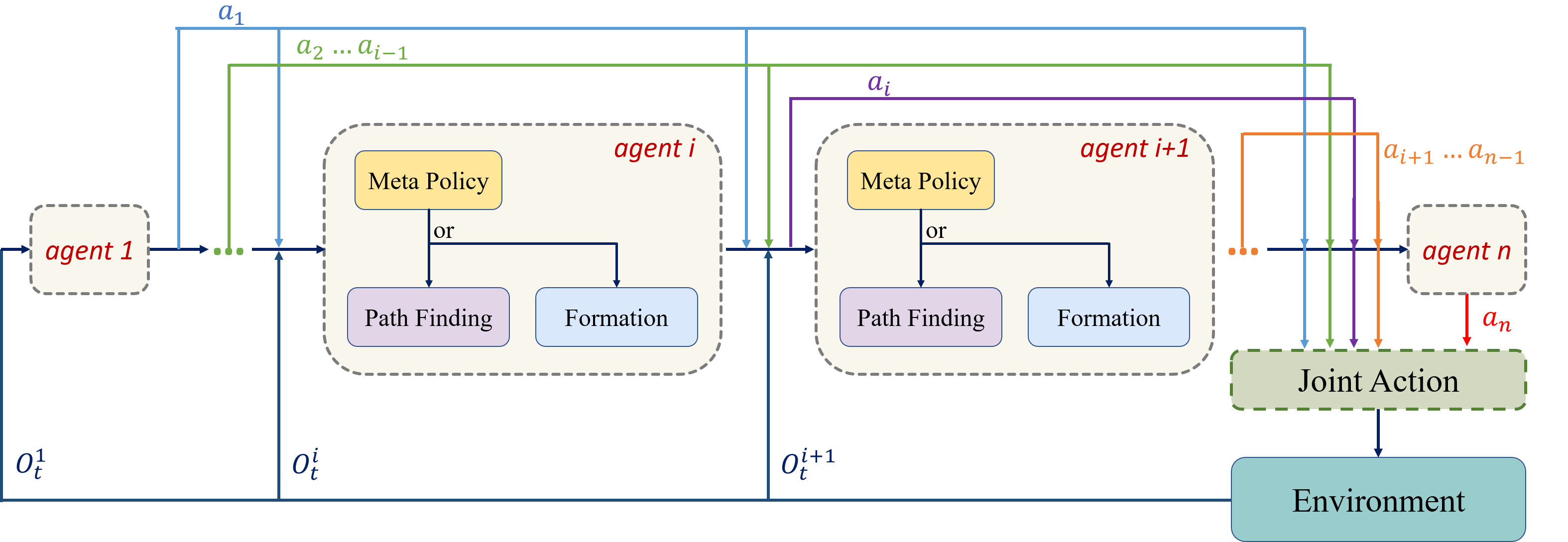}
        \caption{Structure of our overall policy. Agent 1 is the chosen leader in every time step who acts first. Other agents select actions sequentially based on its local observation and all the former agents' actions. Each agent has the same hierarchical structure that uses a meta policy to choose between the path finding policy and formation policy.}
        \label{main}
        \vspace{-15pt}
\end{figure*}
\subsection{Hierarchical Learning}
\label{hrl}
One of the key challenges in training a well-performed policy of MAiF is that the overall task is complex. The MAPF problem itself is hard to learn since it usually takes hundreds of hours to train\cite{primal}. Nevertheless, the formation policy needs to be learned simultaneously. These two goals are conflicted with each other thus the learning process would be unstable. Therefore, learning an end-to-end policy to solve the MAiF problem is difficult. To solve this problem, we propose a novel hierarchical reinforcement learning structure that can decompose the path finding and formation problem into unrelated ones. Then we can train path finding and formation policy separately. By defining unique rewards for each policy, they can learn their own task without being interrupted by optimizing the other task.

However, we have not already solved MAiF problem even if we have a well-trained path finding policy and formation policy. We still need the policy to decide which policy should be used at each time step. Moreover, we can train a meta policy to do so. The meta policy's task is to balance the path finding task and keep formation task by optimizing its own designed reward. As the low policy(path finding and formation policy) is already well-trained, there is no worry about the mutual interference in low policy learning process. However, we still need to design a suitable reward for meta policy to solve multi-objective learning tasks, which will discuss in section \ref{morl}. Nevertheless, as the meta policy has lower dimensions of action and faces a simpler task to learn, optimizing multi-objective tasks can be done much more efficiently.
The overall algorithm can be viewed as Algorithm \ref{algo}.

\begin{algorithm}[h]
        \small
        \caption{Hierarchical learning}
        Initialize $\pi_{meta}(a_m|o_t;\theta_m)$,$\pi_{p}(a_p|o_t;\theta_p)$,$\pi_{f}(a_f|o_t;\theta_f)$

        Initialize three different environments $E_m$,$E_p$,$E_f$

        Pretrain $\pi_{p}$ and $\pi_{f}$ separately in $E_p$,$E_f$

        Initialize replay buffer $R_m$

        Initialize Q-Network $Q_{\theta_m}$ and target Q-Network $Q_{\theta_m'}$
        \begin{algorithmic}[1]

                \For{n episodes = 1 to N}

                \State Agents take meta action $a_m = \pi_m(o)$

                \If{$a_m$ is using path finding policy}

                \State $a_{low} = \pi_p(o)$
                \Else

                \State $a_{low} = \pi_m(o)$
                \EndIf

                \State Store $<o, a_m, r, o'>$ in $R_m$

                \State Sample a mini-batch $B_m$ from $R_m$

                \State Perform a gradient decent step on

                \State $(y-Q_{\theta_m}(o, a))_{B_m}^2$,

                \State where $y=r+\tau Q_{\theta_m'}(o',argmax_{a_m'}(Q_{\theta_m}(o', a_m'))$.

                \If{n mod $I_{TargetUpdate}$}

                \State Update $Q_{\theta_m'}: \theta_m' \gets \theta_m$

                \EndIf

                \EndFor
        \end{algorithmic}
        \label{algo}
\end{algorithm}

\subsection{Multi-Objective Learning}
\label{morl}
In this work, we use a linear scalarization function to define a utility over a vector-valued reward and thereby reducing the dimensionality of the multi-objective reward vector to a single, scalar value.

The linear scalarization function $f$ is a function that projects a vector v to a scalar: $v_w=f(v,w)$, where $w$ is a weight vector parameterizing $f$. In our situation, our meta policy needs to balance the formation reward and path finding reward, so our scalarization function $f$ can be viewed as $v_w=f((r_p,r_f),w)$. $r_p$ and $r_f$ stands for the reward of path finding and formation, in this case, $w$ is a two dimensions vector.

The values of $w$ are usually decided manually in previous works, which is infeasible in our situation as we do not know the accurate upper limits of our rewards. Benefited from the structure of hierarchical reinforcement learning, we introduce a novel MORL algorithm. It calculates a theoretical weight that balances path finding policy and formation policy by making all rewards have equal influence on the final RL value function. Then we choose other values around this basic weight to get the Pareto fronts. Notably, the method can not only use in MAiF problem but also solve other MORL problems. Let $w_f$ be the basic $w$ for $r_f$, $L_{f}$ be the formation loss and $R^*$ be the range, we have

\begin{small}
        \begin{equation}
                w_f = \frac{T}{R^*(\mathbb{E}[ \sum_{t=0}^{T}\Delta L_{f}(o_t)])}
                \label{eq1}
        \end{equation}
\end{small}

$Proof$: Firstly, we can have the optimal value at a state is given by the state-value function
\begin{equation}
        V^{*}\left(o_{t}\right)=\max _{\pi} \mathbb{E}\left[\sum_{t=0} \gamma^{t} f(R\left(o_{t}\right))\right]
        \label{eq2}
\end{equation}
where
\begin{equation*}
        R\left(o_{t}\right)=\max _{\pi} R\left(o_{t}, a_{t}\right)
\end{equation*}

Given a particular set of weights $\mathbf{w}$, we substitute scalarization function $f$ into Eq. \ref{eq2} to obtain
\begin{small}
        \begin{equation}
                \begin{multlined}
                        V^{*}\left(o_{t} \mid \mathbf{w}\right)=\max _{\pi} \mathbb{E}\left[\sum_{t=0}^{\infty} \sum_{a_{t} \in A} \gamma^{t}\left(w_{1} r_{1}\left(o_{t}, a_{t}\right)+\cdots\right.\right. \\
                                \left.\left.+w_{n} r_{n}\left(o_{t}, a_{t}\right)\right)\right]
                \end{multlined}
                \label{eq4}
        \end{equation}
\end{small}

We substitute $w_p$, $w_f$, $r_p$ and $r_f$ into Eq. \ref{eq4} to obtain
\begin{small}
        \begin{equation}
                \begin{multlined}
                        V^{*}\left(o_{t} \mid \mathbf{w}\right)=\max _{\pi} \mathbb{E}\left[\sum_{t=0}^{\infty} \sum_{a_{t} \in A} \gamma^{t}\left(w_{p} r_{p}\left(o_{t}, a_{t}\right) \right.\right.\\
                                \left.\left. +w_{f} r_{f}\left(o_{t}, a_{t}\right)\right)\right]
                \end{multlined}
        \end{equation}
\end{small}

Here, if we want to normalize the $r_p$ and $r_f$, we can let $V^{*}\left(o_{t} \mid \mathbf{w}\right)$ stay the same value if we take optimum $a_t$ whether to max $r_p$ or $r_f$, we note the optimum $a_t^*$ as $a_{p}^*$ and $a_{f}^*$ separately.

\begin{small}
        \begin{equation}
                \begin{multlined}
                        \Delta V^{*}\left(o_{t} \mid \mathbf{w}\right)=
                        \\
                        \mathbb{E}\left[\sum_{t=0}^{\infty} \gamma^{t}\left(w_{p} r_{p}\left(o_{t}, a_{p}^*\right) +w_{f} r_{f}\left(o_{t}, a_{p}^*\right)\right)\right]- \\
                        \mathbb{E}\left[\sum_{t=0}^{\infty} \gamma^{t}\left(w_{p} r_{p}\left(o_{t}, a_{f}^*\right) +w_{f} r_{f}\left(o_{t}, a_{f}^*\right)\right)\right]
                \end{multlined}
                \label{eq6}
        \end{equation}
\end{small}

Since we have already known $r_p(o_t,a_t)$ is 1 only when $a_t$ is $a_p$. Eq. \ref{eq6} can be simplified as
\begin{small}
        \begin{equation}
                \begin{multlined}
                        \Delta V^{*}\left(o_{t} \mid \mathbf{w}\right)=
                        \mathbb{E}\left[\sum_{t=0}^{\infty}  \gamma^{t}\left(1 +w_{f} r_{f}\left(o_{t}, a_{p}^*\right)-w_{f} r_{f}\left(o_{t}, a_{f}^*\right)\right)\right]
                \end{multlined}
        \end{equation}
\end{small}
We can find here that if we want our $\Delta V^{*}\left(o_{t} \mid \mathbf{w}\right)$ equals 0, we can just make
\begin{equation}
        \mathbb{E}\left[\sum_{t=0}^{\infty}  \gamma^{t}\left(1 +w_{f} r_{f}\left(o_{t}, a_{p}^*\right)-w_{f} r_{f}\left(o_{t}, a_{f}^*\right)\right)\right]  = 0
        \label{eq7}
\end{equation}
Eq. \ref{eq7} is only related with reward function $r_f$. And we know $r_f$ stands for formation loss $L_{f}$, we have
\begin{equation}
        r_{f}\left(o_{t}, a_{t}\right) = L_{f}(o_{t+1})
        \label{eq8}
\end{equation}
we substitute Eq. \ref{eq8} into Eq. \ref{eq7} and take $\gamma$ equals 1 for simplify
\begin{equation}
        \mathbb{E}\left[\sum_{t=0}^{\infty} \left(1 +w_{f}\left( L_{f}(o_{t+1}^-)-L_{f}(o_{t+1}^*)\right)\right)\right]  = 0
        \label{eq9}
\end{equation}
in which $L_{f}(o_{t+1}^-)$ means the worst formation loss(comparing to last time step) for time step $o_{t+1}$ and $L_{f}(o_{t+1}^*)$ is the best one. If we define
\begin{equation}
        \Delta L_{f}(o_t) = L_{f}(o_{t+1}) - L_{f}(o_{t})
        \label{eq10}
\end{equation}
We have
\begin{small}
        \begin{equation}
                \centering
                \begin{multlined}
                        \max (\Delta L_{f}(o_t)) = L_{f}(o_{t+1}^*) - L_{f}(o_{t})\\
                        \min (\Delta L_{f}(o_t)) = L_{f}(o_{t+1}^-) - L_{f}(o_{t})\\
                \end{multlined}
        \end{equation}
\end{small}
So,
\begin{small}
        \begin{equation}
                \begin{multlined}
                        \max (\Delta L_{f}(o_t)) - \min (\Delta L_{f}(o_t)) =\\ L_{f}(o_{t+1}^*)-L_{f}(o_{t+1}^-)
                \end{multlined}
                \label{eq12}
        \end{equation}
\end{small}
Then, we substitute Eq. \ref{eq12} into Eq. \ref{eq9} :
\begin{small}
        \begin{equation}
                \begin{multlined}
                        \mathbb{E}\left[\sum_{t=0}^{\infty} \left(1 +w_{f}\left( L_{f}(o_{t+1}^-)-L_{f}(o_{t+1}^*)\right)\right)\right] \\
                        = T + w_{f}( \min\mathbb{E}[ \sum_{t=0}^{T}\Delta L_{f}(o_t)] -  \max\mathbb{E}[ \sum_{t=0}^{T}\Delta L_{f}(o_t)]) = 0
                \end{multlined}
                \label{eq13}
        \end{equation}
\end{small}

Here, we transfer the problem to estimate the expectation of $\sum_{t=0}^{T}\Delta L_{f}(o_t)$. This can be easily estimated by evaluating a well-trained formation policy and a random policy, let $R^*$ stands for range here.

\begin{small}
        \begin{equation}
                w_f = \frac{T}{R^*(\mathbb{E}[ \sum_{t=0}^{T}\Delta L_{f}(o_t)])}
                \label{eq14}
        \end{equation}
\end{small}

\begin{table*}[t]
        \centering
        \caption{RESULTS OVER DIFFERENT EXPERIMENT SETTINGS.}
        \resizebox{0.95\linewidth}{!}{
                \begin{tabular}{|c|c|c|c|c|c|c|c|c|c|c|c|c|c|c|c|c|c|c|}
                        \hline
                        \multicolumn{3}{|c|} { Environment Setting } &
                        \multicolumn{4}{c|} {Makespan}               &
                        \multicolumn{4}{c|} {Formation Loss}         &
                        \multicolumn{4}{c|} {Success Rate}           &
                        \multicolumn{4}{c|} {Runtime(s)}                                                                                                                                                                                         \\
                        \hline
                        map size                                     & agent & d    & Ours              & CBS  & SW   & A* & Ours            & CBS  & SW   & A*  & Ours             & CBS & SW  & A*  & Ours            & CBS    & SW    & A*    \\
                        \hline
                        $20\times20$                                 & 3     & 0.15 & 41.3              & 34   & 35.4 & -  & 0.5             & 3.37 & 0.12 & -   & $\mathbf{1 . 0}$ & 1.0 & 1.0 & 0.0 & 0.27            & 0.002  & 0.05  & -     \\
                        \hline
                        $20\times20$                                 & 3     & 0.05 & 34.2              & 34   & 34   & 34 & 0.01            & 0.2  & 0.0  & 0.0 & $\mathbf{1 . 0}$ & 1.0 & 1.0 & 1.0 & 0.24            & 0.0002 & 0.002 & 0.003 \\
                        \hline
                        $20\times20$                                 & 5     & 0.05 & $\mathbf{30}$     & 30   & 30   & 30 & $\mathbf{0.0}$  & 1.05 & 0.0  & 0.0 & 0.8              & 1.0 & 1.0 & 1.0 & 0.39            & 0.0004 & 0.002 & 0.003 \\
                        \hline
                        $512\times512$                               & 3     & 0.15 & 1131.2            & 1018 & -    & -  & $\mathbf{0.37}$ & 9.26 & -    & -   & $\mathbf{1 . 0}$ & 1.0 & 0.0 & 0.0 & $\mathbf{8.47}$ & 171.58 & -     & -     \\
                        \hline
                        $512\times512$                               & 4     & 0.05 & 1713.6            & 1014 & 1014 & -  & 1.07            & 10.4 & 0.0  & -   & $\mathbf{1 . 0}$ & 1.0 & 1.0 & 0.0 & 16.3            & 58.9   & 4.5   & -     \\
                        \hline
                        $1024\times1024$                             & 3     & 0.05 & 2378.7            & 2042 & -    & -  & $\mathbf{0.17}$ & 24.7 & -    & -   & $\mathbf{1 . 0}$ & 1.0 & 0.0 & 0.0 & $\mathbf{18.1}$ & $>$300 & -     & -     \\
                        \hline
                        $1024\times1024$                             & 3     & 0.15 & $\mathbf{2215.5}$ & -    & -    & -  & $\mathbf{0.38}$ & -    & -    & -   & $\mathbf{1 . 0}$ & 0.0 & 0.0 & 0.0 & $\mathbf{17.2}$ & -      & -     & -     \\
                        \hline
                \end{tabular}}
        \label{table2}
        \vspace{-10pt}
\end{table*}

\subsection{Decentralized Cooperative Multi-Agent Learning}
It is challenging to train a fully decentralized multi-agent cooperative policy, especially when the task needs highly cooperative agents. We use VDN\cite{vdn} to train agents in a centralized way but can act decentralized. However, cooperative policies can not perform well when it relies on a single agent's observation. To solve this problem, we propose a novel communication method based on theory of mind. The theory of mind indicates that agents can interpret others' actions and act in a more informative way. Thus, we use a combination of single agent observation and other agents' joint actions $u_t^-$ as the input for all policies. Our communication method only involves low-dimension action information, which takes up little bandwidth and makes it suitable to deploy in the real world.

However, if every agent's action is based on other agents' actions, we can not calculate all actions at the same time step. In other words, one agent must move first without inferring other agents' actions. Then other agents can select movements based on the performed actions of former ones. In such a situation, the joint action is significantly influenced by the first taken action because all other agents cooperate with the former ones. So how to choose the first move agent, called leader, is essential for improving policy performance.

In our work, we propose a dynamic leader chosen method. When executing formation policy, we choose the leader as the agent in the middle of the formation. The reason is that the middle agent has the broadest view of all agents' relative positions. It can move to an optimum position in favor of resuming or keeping formation. While executing the path finding policy, we choose the agent in the front as the leader, for that the leader can observe the future path and choose a path with fewer obstacles to go.

\begin{figure}[t]
        \centering
        \includegraphics[width=0.95\linewidth]{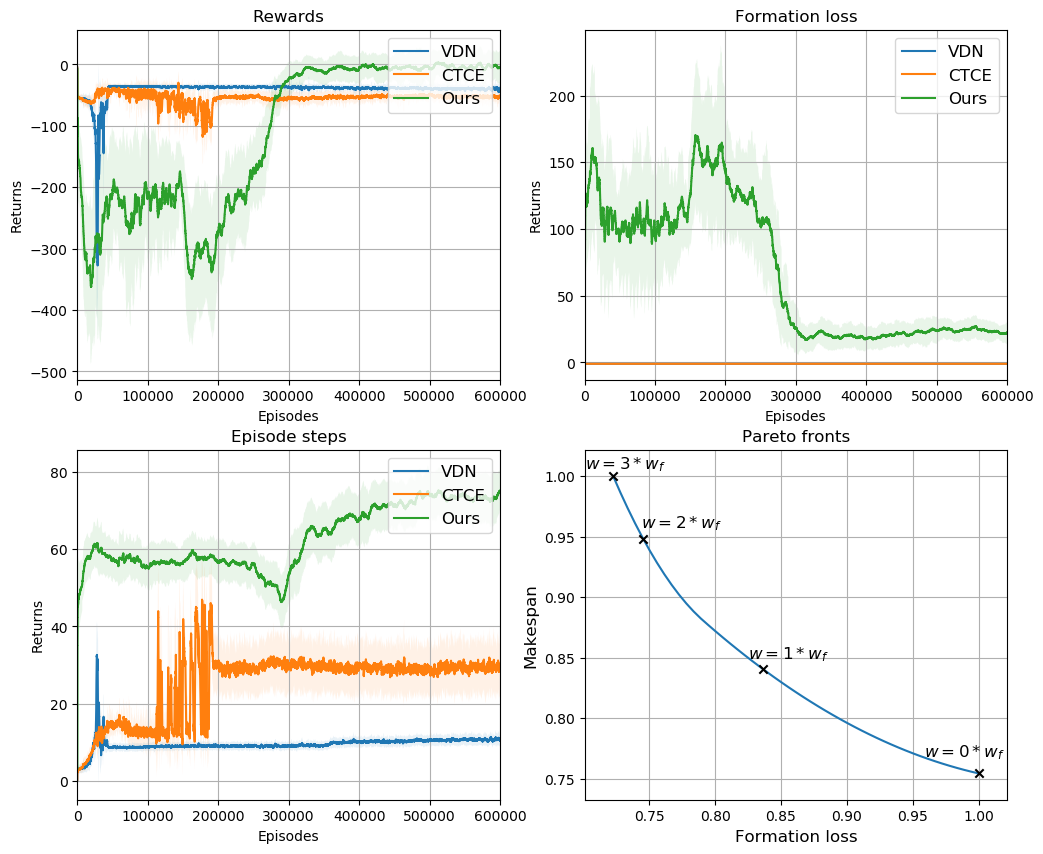}
        \caption{Left: rewards(top) and episode steps(bottom) of all RL methods. Right: formation loss(top) of all RL method and Pareto fronts(bottom) of our method. For fair comparing, we train both VDN and CTCE three times longer than our method as we have pre-trained path finding and formation policies. In order to facilitate the display, we have scaled our method on the horizontal axis. We also normalize the coordinate axis of the Pareto fronts.}
        \label{rl}
\end{figure}

\section{EXPERIMENTS}
\subsection{Experiment Settings}
For our experiments, we chose our three centralized planning methods for comparison: Joint-State A*\cite{felner2017search}, SWARM-MAPF\cite{li2020moving} and  Conflict-Based Search (CBS)\cite{sharon2015conflict}. The joint-state A* performs poorly on scalability, yet it finds the optimal Pareto fronts. SWARM-MAPF is the state-of-the-art centralized planner which gives a nearly optimal solution to MAiF. CBS can give us a baseline as the pure MAPF method that does not optimize formation loss. For all centralized planning methods except CBS, we used a time limit of 300s. Note that all other centralized methods have access to the whole state of the environment. In contrast, our method assumes that each agent only has partial observability of the environment and plans as an individual. As for other RL methods, to the best of our knowledge, none of the previous work in RL can directly apply to MAiF. So, we compare with VDN\cite{vdn} and a centralized training centralized executing(CTCE) method as RL baseline methods. These two methods also use action clipping and the basic $w_f$ to balance rewards.

In our experiments, we compare makespan, formation loss, runtime and success rate with centralized planning methods and RL methods.
For a fair comparison, our method is tested in untrained maps. The test maps are randomly generated and have various sizes and obstacle densities.

For all decentralized partially observable methods, we add a few random steps in each testing episode to try different paths that can find the nearly optimum solution as the final performance. All experiments are carried out on the same computer, equipped with an Intel i7-7700K, 16GB RAM and an NVIDIA GTX1080Ti.

Our benchmark and code will be released in \url{https://github.com/zijinoier/mater}.

\subsection{Training Details}
\subsubsection{Environment}
We apply a grid world simulation environment, just as Fig. \ref{obs} shows. The map size is $\{20, 32, 512, 1024\}$. The obstacle density is $\{0.05, 0.15\}$. We make a limit length of walls as half of the length of the agents' view field. This can prevent the agent from being completely separated in the field of view. For each map, the top-left $5\times5$ or $10\times10$ cells (depends on map size) are possible start locations, and the bottom-right $5\times5$ or $10\times10$ cells are possible goal locations. The formation is placed in start locations. During the training process, the environment maps are randomly selected at the beginning of each episode in a map pool with 100 different maps. The map size is 32 and the obstacle density is 0.15. During the testing process, the maps are generated randomly in each episode and tested ten times.

\subsubsection{Parameters}: We use a discount factor $\gamma$ of 0.95, an episode length of 3 times of map size(maximum), and a batch size of 32. We use basic $w_f$ to balance the reward. All policies use the same network consists of two LSTM layers with 256 units. We use the Adam optimizer with a learning rate $2\cdot10^{-6}$.


\subsection{Result}

Table. \ref{table2} shows the results of our method comparing with centralized methods. All experiments are evaluated in 10 different maps. The formation loss value is normalized by map size. Based on our results, we show that our method is transferable and can be adapted to any map size and obstacle density. Furthermore, we notice that our approach performs extremely well in the large scale world that other centralized methods can not handle. The reason is that our method acts only refer to local observation and the planning time grows linearly as the map size or agent number grows. Our method can handle different obstacle densities without paying extra computing expenses. Meanwhile, the centralized methods' runtime grows exponentially as the map size, agent number or obstacle density grows. Therefore, they cannot cope with maps of large size or high obstacle density. In small-sized maps, our method can achieve performance similar to centralized methods, even if our method plans separately based on a limited field of view.

We present our method's result compared with end-to-end RL methods and the Pareto fronts of our method in Fig. \ref{rl}. We notice that other end-to-end RL methods can hardly get out of a ground performance. Both of them can not even learn to reach the goal position, so we can not compare formation loss with them. Even evaluated in episode steps, we observe that CTCE methods are stuck in some kind of local optimal policy while VDN does not learn anything. The reason is that they want to learn both path finding task and formation task together. Moreover, optimizing two conflict objectives can cause learn in a dilemma. We also notice that the CTCE policy has a better performance than VDN. This is due to the fact that its agents can obtain information from other agents while selecting actions. This information can improve the performance of cooperative policy in the same way as our communication method. Finally, we present our Pareto fronts. We notice that our method reaches a policy that can balance path finding and keeping in formation when the weight of formation loss equals our base weight. This is in line with our theoretical calculations. We can get the whole Pareto fronts by using n times base weight (e.g. 0,2,3 in the figure).


\begin{figure}[t]
        \centering
        \includegraphics[width=0.9\linewidth]{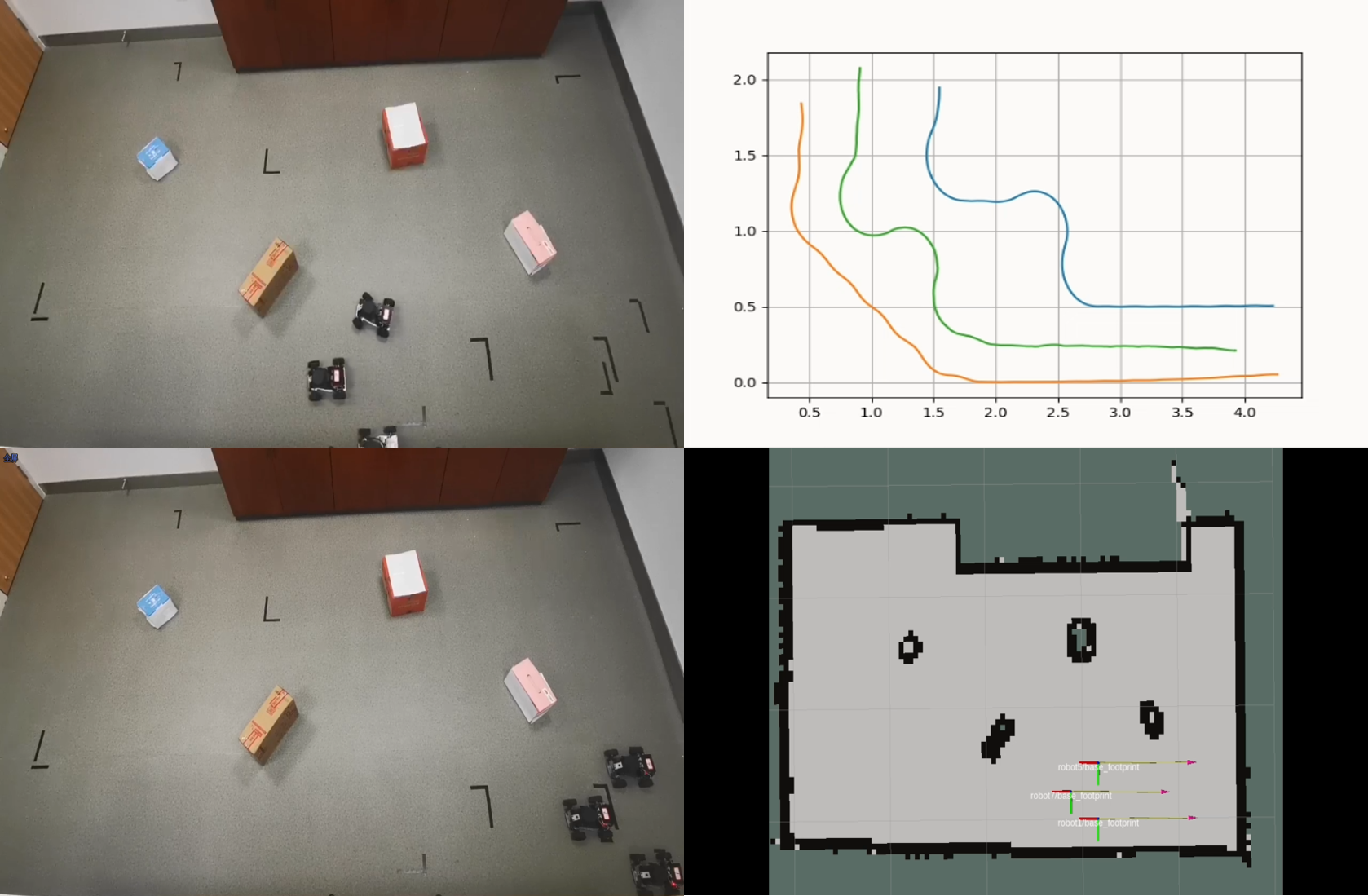}
        \caption{Snapshot of hardware experiment. Left: moving agents(top) and agents' start positions(bottom). Right: trajectories(top) and RVIZ view(bottom) of all agents.}
        \label{hex}
\end{figure}

\subsection{Hardware Experiments}

We also implement our method on a small fleet of Ackermann cars. Due to venue restrictions, we choose an indoor room to simulate a three cars formation. Each car plans on the host computer using our decentralized approach. And we transform the discrete action into continual action so the Ackermann cars can carry out. The result indicates that our method has clear sim-to-real capabilities, as the planning time per step is below 0.1s on a laptop computer. Fig. \ref{hex} shows our Ackermann cars and experiment environment.

\section{CONCLUSIONS}
In this paper, we present a new decentralized partially observable approach to multi-agent in formation. It utilizes a novel hierarchical reinforcement learning structure that can solve multi-objective reinforcement learning problems effectively. Furthermore, we propose a theoretical weight calculation method that makes every task's reward has equal influence on the final RL value function. Additionally, we introduce a communication method that helps agents cooperate with each other. Through an extensive set of experiments, we show that our method outperforms several end-to-end RL algorithms and can scale to various formations, world sizes and obstacle densities. Our method performs well in large-scale worlds where centralized methods struggle. Finally, we present a demonstration where we deploy our method in real-world robots, showing our method sim-to-real ability. However, our method still needs to retrain the formation policy when the formation changes. Our future work will focus on finding a general formation policy that can deal with multiple formations.



\bibliographystyle{IEEEtran}
\bibliography{reference}

\end{document}